\documentclass[conference]{IEEEtran}
\IEEEoverridecommandlockouts
\usepackage{cite}
\usepackage{amsmath,amssymb,amsfonts}
\usepackage{algorithmic}
\usepackage{graphicx}
\usepackage{textcomp}
\usepackage{xcolor}
\def\BibTeX{{\rm B\kern-.05em{\sc i\kern-.025em b}\kern-.08em
    T\kern-.1667em\lower.7ex\hbox{E}\kern-.125emX}}
    
\usepackage{tikz}
\usepackage{caption}
\usepackage{subcaption}
\usetikzlibrary{spy,calc}
\usepackage{verbatim}
\usepackage{booktabs,array} 
\usepackage{url}

\usepackage{siunitx}
\usepackage{microtype}
\usepackage{cite}
\usepackage{amsmath,amsfonts,amssymb,amsthm}
\usepackage{mathtools}

\usepackage[ruled,vlined]{algorithm2e}
\usepackage{nicefrac}

\usepackage{pgfplots}
\pgfplotsset{compat=newest, legend style={at={(1,0.05)},anchor=south east}
}

\usepackage{hyperref}

\hypersetup{hidelinks,
backref=true,
pagebackref=true,
hyperindex=true,
breaklinks=true,
colorlinks=true,
urlcolor=blue,
bookmarks=true}

\usepackage{cleveref}

\usepackage{subcaption}

\crefname{section}{Sec.}{Sections}
\crefname{figure}{Fig.}{Figure}
\crefname{table}{Tab.}{Table}
\crefname{equation}{Equ.}{Equation}

\newcolumntype{L}[1]{>{\raggedright\arraybackslash}p{#1}}

\makeatletter
\DeclareRobustCommand\onedot{\futurelet\@let@token\@onedot}
\def\@onedot{\ifx\@let@token.\else.\null\fi\xspace}
\makeatother

\newcommand{\etal}[1]{#1~et~al\onedot}

\newcommand{\eg}{e.\,g.,\xspace}

\newcommand{\cf}{cf\onedot}
\newcommand{\ie}{i.\,e.,\xspace}

\makeatletter
\newcommand{\thickhline}{%
    \noalign {\ifnum 0=`}\fi \hrule height 1pt
    \futurelet \reserved@a \@xhline
}
\usepackage{datetime}

\newcommand{\retweet}{\textsc{ReTweet}\xspace}

\makeatletter
\renewcommand\footnoterule{%
  \kern-3\p@
  \hrule\@width.4\columnwidth
  \kern2.6\p@}
  \makeatother

\begin{document}

\title{How Will Your Tweet Be Received? Predicting the Sentiment Polarity of Tweet Replies\thanks{$^{*}$~\textbf{\textit{Corresponding author}}.}
\thanks{A major part of the research was carried out by the first two authors.}}

\author{\IEEEauthorblockN{Soroosh Tayebi Arasteh$^{*}$\IEEEauthorrefmark{2}\IEEEauthorrefmark{3},
Mehrpad Monajem\IEEEauthorrefmark{2},
Vincent Christlein\IEEEauthorrefmark{2}, 
Philipp Heinrich\IEEEauthorrefmark{2}, 
\\
Anguelos Nicolaou\IEEEauthorrefmark{2}, 
Hamidreza Naderi Boldaji\IEEEauthorrefmark{2}, 
Mahshad Lotfinia\IEEEauthorrefmark{4} and
Stefan Evert\IEEEauthorrefmark{2}}

\\
\IEEEauthorblockA{\IEEEauthorrefmark{2}Friedrich-Alexander-Universität Erlangen-Nürnberg, Germany}
\IEEEauthorblockA{\IEEEauthorrefmark{3}Harvard Medical School, United States}
\IEEEauthorblockA{\IEEEauthorrefmark{4}Sharif University of Technology, Iran}
soroosh.arasteh@fau.de}

\maketitle

\begin{abstract}
Twitter sentiment analysis, which often focuses on predicting the polarity of tweets, has attracted increasing attention over the last years, in particular with the rise of deep learning (DL).
In this paper, we propose a new task: predicting the predominant sentiment among (first-order) replies to a given tweet. Therefore, we created \retweet, a large dataset of tweets and replies manually annotated with sentiment labels.
As a strong baseline, we propose a two-stage DL-based method:
first, we create automatically labeled training data by applying a standard sentiment classifier to tweet replies and aggregating its predictions for each original tweet; our rationale is that individual errors made by the classifier are likely to cancel out in the aggregation step.
Second, we use the automatically labeled data for supervised training of a neural network to predict reply sentiment from the original tweets. The resulting classifier is evaluated on the new \retweet dataset, showing promising results, especially considering that it has been trained without any manually labeled data.
Both the dataset and the baseline implementation are publicly available.

\end{abstract}


\section{Introduction}

Twitter is one of the dominant social media platforms and has become a major source for local and global news as well as political argumentation. Sentiment analysis is a research area in Natural Language Processing that aims to identify the opinions, attitudes or emotions expressed in a text document or sentence, often with respect to a particular topic~\cite{liu2012sentiment}.
Twitter sentiment analysis usually involves detecting whether a given tweet expresses a \emph{positive}, \emph{negative}, or \emph{neutral} sentiment polarity, but has also looked at subparts of the tweet (words, phrases) or at aggregated sentiment in tweet collections, as well as sentiment towards a specific topic, \eg a person, product, or event \cite{rosenthal-etal-2017-semeval}.

\etal{Nakov}~\cite{nakov-etal-2013-semeval} first introduced the task of message- and expression-level sentiment analysis on tweets;
since then many further challenges have emerged.
\etal{Rosenthal}~\cite{rosenthal2019semeval2015} introduced the task of sentiment towards a topic. Variations of the task include more fine-grained classification (\eg a five-point sentiment classification with labels \emph{highly positive, positive, neutral, negative, highly negative}) and quantification of the overall distribution towards a given topic~\cite{nakov-etal-2016-semeval}. There is also work on
sentiment analysis of figurative language \cite{ghosh-etal-2015-semeval} and detecting stance in tweets~\cite{mohammad-etal-2016-semeval}.
Other tasks such as out-of-context sentiment intensity of words and phrases~\cite{kiritchenko-etal-2016-semeval} and implicit event polarity~\cite{russo-etal-2015-semeval} are closely related, but do not target tweets.

In this paper, we introduce a new challenge in Twitter sentiment analysis: given only an original tweet, predict the predominant polarity among all first-order replies this source tweet will receive, \ie whether the audience will overall react positively, negatively, or neutrally to the tweet. It is challenging to collect large amounts of training data for this task because, for each training item, annotators need to read all first-order replies in order to label the source tweet. 
Therefore, we propose to 
generate automatically labeled training data by applying a standard Twitter sentiment classifier to sets of replies and aggregating the results. These data can then be used for standard supervised training of the final classifier.
For evaluation purposes, we have created \retweet, a publicly available\footnote{Accessible under this link: \url{https://www.kaggle.com/soroosharasteh/retweet/}.}
test set with manual annotation, and use it for validating our proposed method\footnote{The source code is accessible below:
\url{https://github.com/starasteh/retweet/}.}.


\section{Standard Tweet Sentiment Classification}
\label{section:tweet}

For our method presented in \cref{section:Methodology}, we need a state-of-the-art sentiment classifier for tweets. As a baseline, we evaluate models for task 10.B of SemEval 2015~\cite{rosenthal2019semeval2015}, which is the task of three-way classification according to \emph{positive}, \emph{negative}, or \emph{neutral} sentiment for English tweets. We use \num{51875} manually labeled tweets from the SemEval datasets~\cite{nakov-etal-2013-semeval,rosenthal2019semeval2014,rosenthal2019semeval2015,nakov-etal-2016-semeval,rosenthal-etal-2017-semeval} as training data and \num{9155} tweets as validation data.
We combined individual training and test sets to ensure there is no overlap.

Our own classifier is based on a recurrent neural network. The tweets are tokenized using SpaCy~\cite{honnibal-johnson-2015-improved}. The embedding layer uses a Vocabulary of the 50K the most frequent words in the training data; out-of-vocabulary words are replaced with a special ``unknown'' token (\texttt{<unk>}). The embeddings are initialised with a 200-dimensional GloVe~\cite{Pennington14glove:global} model trained on 27 billion tweets.

\begin{figure}[t]
    \centering
        \includegraphics[scale=0.4]{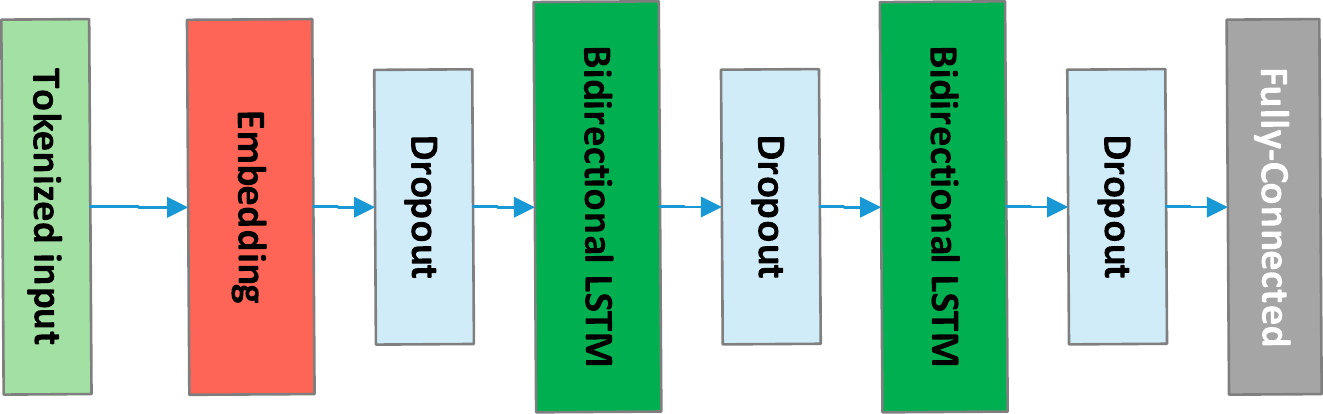}
    \caption{The network architecture.}
    \label{fig:architecture}
\end{figure}
The main building-blocks of our model are two Bi-directional Long-Short Term Memory Units (BiLSTM)~\cite{hochreiter1997long} with the hidden and cell states of length 256. First of all, a dropout layer is applied to the input embedding vectors. Then the result is fed into a BiLSTM followed by another dropout layer and then the second BiLSTM followed by another dropout layer. Finally, a fully connected layer (followed by a SoftMax) is used for the classification.
\cref{fig:architecture} shows our selected architecture. 
All dropout layers use a drop probability of 0.5.
As loss, we chose the weighted cross-entropy with inverted class frequencies of the training data as loss weights to counteract the imbalanced dataset. The model is optimized using the Adam optimizer, learning rate of $10^{-4}$ and weight decay of $10^{-5}$.

To compare our tweet classifier with other state-of-the-art models, we test our trained model on the 2014 and 2015 test sets of SemEval, which consist of \num{1852} and \num{2389} manually labeled tweets, respectively. In order to be consistent with historical editions of this competition, we use the average $F_1$ score of the positive and negative classes as the metric of interest, see \cref{eq:one}. Note that, except accuracy, which is calculated on the whole set, the rest of the evaluation scores are all calculated according to \cref{eq:one} throughout this paper, \ie discarding the neutral class results:
\begin{equation}
 \text{score} =  \frac{\text{score}_\text{pos} + \text{score}_\text{neg}}{2}
\label{eq:one}
\end{equation}

\Cref{tab:semval_results} shows that our tweet classifier is in the range of state-of-the-art and conventional machine learning models and we can proceed to utilize it for automatic labeling. 

\begin{table}[t]
\centering
\caption{Evaluation of sentiment classifiers on the SemEval 2014 and 2015 official test sets.
The scores present the average F1 scores of the positive and negative classes according to \cref{eq:one}. 
Numbers taken from~\protect\cite{cliche-2017-bb,rosenthal2019semeval2014,rosenthal2019semeval2015}.}
\label{tab:semval_results}
\begin{tabular}{lcc} 
\toprule
\textbf{System} & \textbf{2014} & \textbf{2015} \\
\midrule
Logistic regression on 1-3 grams baseline & 0.629 & 0.586 \\
The 9th place of the original task & 0.674 & 0.620 \\
The winner of the original task & 0.709 & 0.648 \\
The state-of-the-art & 0.748 & 0.688 \\
\emph{Our sentiment polarity classifier} (\cf\cref{section:tweet}) & \textbf{0.652} & \textbf{0.624} \\
\bottomrule
\end{tabular}
\end{table}


\section{Methodology}
\label{section:Methodology}

Having a trained sentiment classifier, we can start solving the main problem of predicting overall sentiment polarity of the replies.
In the automatic labeling process, given a source tweet and a set of corresponding replies, the trained sentiment classifier is first applied to all the replies as inputs, while ignoring the source tweet. Consequently, the network predicts a sentiment category for each reply, which are collected in a vector.
A heuristic algorithm (\cf\cref{alg:algo}) is used to derive a final label for the overall sentiment of replies to the source tweet based on this label vector.
\begin{algorithm}[t]
\KwResult{One label for every tweet based on its replies.}
 \For{all tweets}{
  \uIf{\#total neutral replies $>$ \SI{85}{\percent} of the total replies}{
    Label = neutral \;
  }
  \Else{
  \uIf{\#total positives replies $>$ \#\,$1.5\times$ of total negative replies}{
    Label = positive \;
  }
  \uElseIf{\#total negative replies $>$ \#\,$1.6\times$ of total positives replies}{
    Label = negative \;
  }
  \Else{
    Label = neutral \;
  }
  }}
\caption{Automatic label assignment strategy.}
\label{alg:algo}
\end{algorithm}
Some justifications are needed to support our method, as there is no theoretical motivation or empirical evidence behind the chosen ratios. 

Firstly, we generally expect most of the polarity predictions in the vector to be neutral as not all replies can be expected to carry positive/negative sentiment. Therefore, the overall reply sentiment is only considered to be neutral if the proportion of neutral replies exceeds a fairly high threshold, here set to \SI{85}{\percent}.
Secondly, we observe that usually
every tweet triggers both positive and negative replies as there are different people with different mindsets commenting on the tweet. In other words: there is usually no exclusively positive or negative reply vector for a tweet, even for seemingly clear-cut cases such as, \eg birthday wishes. We therefore assign positive overall sentiment if there are at least $1.5$ times as many positive replies as negative replies, and vice versa.
Note that the exact values were fine-tuned based on various observations on the behavior of the network in different situations.
Finally, we assign \emph{neutral} to a tweet when the total numbers of positive and negative replies were close to each other, as none of them is dominant and there are two sets of contradicting replies, making it difficult to opt for one side.\footnote{Note that despite the name the \emph{neutral} class thus also contains tweets that trigger both positive and negative replies.}


\section{\retweet: Dataset of Tweets \& Overall Predominant Sentiment of their Replies}
\label{section:data}
As there is no publicly available dataset for our purposes, we downloaded data using the Twitter API. We collected \num{35072} training tweets together with a total of \num{1519504} replies, and used \SI{10}{\percent} of them for validation.
To download all of the replies to a tweet, the Search API could be used. However, it is limited to \num{75000} requests per hour, which causes the mining and downloading process to be slow.
Furthermore, using the Twitter API, there is no possibility of obtaining completely random data. Therefore, we tried to make the procedure as random as possible by utilizing two different strategies for data selection, using the collected data in an intermixed manner.

Our first strategy is based on a sample of English tweets obtained by filtering the Twitter stream via a list of cultural keywords~\cite{bennett2005new}. This list consists of 147 words that are deemed to play a ``pivotal role in discussions of culture and society''~\cite[xvii]{bennett2005new}, covering diverse words such as \emph{aesthetics}, \emph{environment}, \emph{feminism}, \emph{power}, or \emph{tourism}. We extracted all tweets in 2019 having a minimum of 20 first-order replies in the dataset. The data come with an obvious caveat: both the source tweet and replies must contain at least one word from the keyword list.\footnote{This makes it highly unlikely that the list of replies for any given source is exhaustive, \ie there will usually be many other first-order replies to the source tweet that are not included in the dataset.} 

As our second approach, we use the GetOldTweets3~\cite{GetOldTweets3} library to download all the replies corresponding to every tweet. To increase randomness, instead of using the same list of keywords as in the first strategy, we manually selected keywords that are likely to trigger long discussions, such as \emph{Coronavirus} and \emph{football}, or strong opinions, such as \emph{birthday}, \emph{war}, or \emph{racism}. In addition to the constraint that every tweet must have at least 20 first-order replies, 
we only accepted tweets (both source tweets and replies) consisting of at least 20 tokens. This is because our tweet classifier (\cf\cref{section:tweet}) is optimized based on the message-level classification paradigm and thus relies on a sufficient number of words in the message.

A total of \num{5015} tweets with all of their corresponding first-order replies, collected by a combination of the two collection strategies, were given to three trained annotators. 
Annotators were asked to judge intuitively whether the replies taken together indicate an overall positive, negative or neutral reaction to the tweet and decide on \emph{one} final sentiment label for the replies.
Annotators were not shown the respective source tweet in order to avoid a bias based on their own reaction to the tweet.
\emph{Positive} and \emph{negative} polarity is defined in the same way as for the SemEval tasks.
In addition to tweets which actually have a neutral overall polarity among their first-order replies, annotators were asked to also assign neutral label to a tweet when neither positive nor negative replies have predominance on the other one.
For the final dataset, we only chose tweets for which all annotators judged unanimously, leading to a data set of \num{1519} source tweets manually labeled with the overall sentiment of their replies (excluding sentiment of the source tweet itself). 

During the manual annotation process, we observed that most disagreements between the annotators were related to neutral labels, \ie either one annotator classified the overall label of tweet as neutral and the other two not, or two classified as neutral and the other one not. Although the annotators were trained in a same way, the concept of neutral label was still not as clear as positive/negative to them. This can be because human brain, when trying to annotate such phrases, expects more to face phrases with positive or negative polarities than neutral and is more sensitive towards these polarities. Thus, chance of having different opinions between negative and positive is very lower than between neutral and positive/negative.
Label distribution in our final test set is as follows: \SI{32.5}{\percent} positive, \SI{37.5}{\percent} negative, and \SI{30.0}{\percent} neutral; the training set contains \SI{23.5}{\percent} positive, \SI{32.5}{\percent} negative, and \SI{44.0}{\percent} neutral labels.

\section{Experiments and Results on \retweet}
\label{section:Results}
We utilize the same data pre-processing, training process, and architecture as for the standard Twitter sentiment classifier (\cref{section:tweet}); the only differences being that the lengths of hidden and cell states are 300, vocabulary is built choosing 750K of the most common words of the training data, and a learning rate of $9\times10^{-5}$ with a decay of $10^{-4}$ is chosen.

\cref{tab:all_results} shows the confusion matrix and evaluation results of the final classifier on the \retweet test set. 
Even though the standard sentiment polarity classifier  (\cref{section:tweet}) used for automatic labelling of the training data (\cref{section:Methodology}), our classifier for response sentiment prediction already achieves highly promising results, especially
considering the general difficulty of message-level sentiment analysis and the indirect nature of the \retweet task.

In order to explore the correlation between tweet sentiments and their reply sentiments, we additionally create a ``direct'' baseline classifier by predicting the sentiment of the original tweet and assuming that its replies will have the same predominant sentiment.
\cref{tab:part1vs2} shows the evaluation metrics of our proposed system versus the ``direct'' baseline on the \retweet test set.  We observe a highly significant increase in $F_1$ score by $15.4$ percent points.

\begin{table}[t]
\centering
\subcaptionbox{\label{tab:results}}{
\begin{tabular}[b]{cccc} \hline
\toprule
\textbf{Accuracy} & \textbf{F1 score} & \textbf{Recall} & \textbf{Precision} \\
\midrule
61.7 & 71.9 & 79.1 & 66.1 \\
\bottomrule
\end{tabular}
}
\qquad
\subcaptionbox{\label{fig:confusionmatrix}}{
\centering
\includegraphics[scale=0.45]{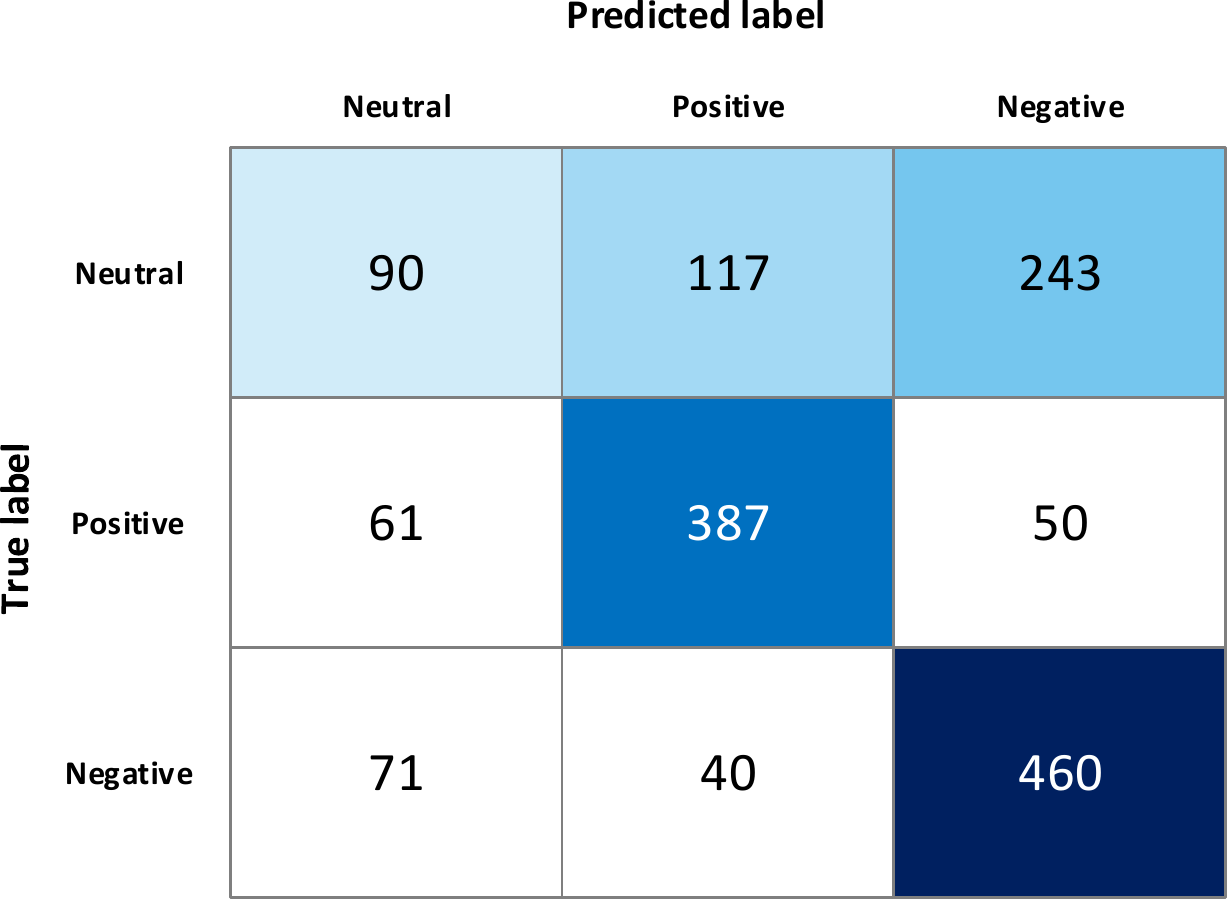}
}
\caption{\subref{tab:results} evaluation results of the final classifier on the \retweet test set. F1 score, recall, and precision are calculated based on \cref{eq:one}. All results given in percent. \subref{fig:confusionmatrix} Detailed confusion matrix.}
\label{tab:all_results}
\end{table}

Furthermore,
we follow the state-of-the-art model of standard tweet sentiment polarity classification proposed by Cliche~\cite{cliche-2017-bb},
implementing a Convolutional Neural Network-based model and averaging its predictions with those of our BiLSTM model.

Firstly, each input embedding vector is fed to three 1D convolutional layers with filter sizes of 3, 4, and 5, respectively, and the Rectified Linear Unit (ReLU) is used as activation function. All filters have the same output feature map dimension of 200. Moreover, in order to remove the dependency of the feature maps on the length, we additionally feed them to 1D max pooling layers with kernel sizes of equal to sentence lengths of each activation. Finally, the results are concatenated, leading to a 600-dimensional feature map for every input, followed by a dropout, and a fully connected layer (followed by a SoftMax)
is used for the classification.

We train the CNN independently with similar optimization and data pre-processing parameters as for our BiLSTM model. 
The evaluation results on the \retweet test set (see \cref{tab:ensemble}), unexpectedly, show only a very slight improvement over the BiLSTM model, which indicates the fact that having already a good enough classifier architecture, the choice of the label assignment algorithm plays a more vital role than maximizing the capacity of the classifier architecture in this task.

Furthermore,
\cref{fig:confusionmatrix} shows that most of the wrong classification results have actually neutral reply polarity. 
It is because, using \cref{alg:algo}, we make our system sensitive to positive/negative sentiments in order to boost our performance of interest. 
And it does not hurt us to lose to a certain extent the neutral prediction performance as long as we are improving in our primary goal.

\begin{table}[t]
\centering
\caption{Comparison between our proposed method (\cref{section:Methodology}) and directly classifying the tweet and assuming its replies will have the same dominant polarity as itself. F1 score, recall, and precision are calculated based on \cref{eq:one}. All results given in percent.}
\label{tab:part1vs2}
\begin{tabular}[b]{ccc} \hline
\toprule
\textbf{Metric} & \textbf{Direct} & \textbf{Proposed} \\
\midrule
Accuracy & 49.7 & 61.7 \\
F1 score & 56.5 & 71.9 \\
Recall & 54.0 & 79.1 \\
Precision & 60.0 & 66.1 \\
\bottomrule
\end{tabular}
\end{table}

\begin{table}[t]
\centering
\caption{Evaluation results of the ensemble model (\cf\cref{section:Results}) on the \retweet test set. F1 score, recall, and precision are calculated based on \cref{eq:one}. All results given in percent.}
\label{tab:ensemble}
\begin{tabular}{cccc} 
\toprule
\textbf{Accuracy} & \textbf{F1 score} & \textbf{Recall} & \textbf{Precision} \\
\midrule
62.0 & 73.2 & 81.0 & 66.8 \\
\bottomrule
\end{tabular}
\end{table}

\section{Conclusion And Discussion}
\label{section:Discussion}
In this paper, we introduced a new challenge: prediction of the overall polarity of first-order replies to an English source tweet. As a baseline for solving this task, we proposed a Deep Learning approach. 
We first predicted the overall message-level polarity of the tweets, \ie whether it is received positive, negative, or neutral, which led to creation of automatic labels for replies, and subsequently trained a network that predicts the predominant reaction of the tweet audience. 
We used a heuristic algorithm for the final label selection in the automatic labeling process.

Additionally, we created \retweet, the first public dataset for sentiment prediction of first-order Twitter replies. 
The evaluation results of our approach on \retweet show its effectiveness.
Although utilizing more training data may still somewhat increase the performance, whether a reply is positive or negative, except for extreme cases, may not purely be determined only by the content of the tweet itself but in a rather dynamic environment. Therefore, our proposed method serves as an upper-bound baseline in prediction of the polarity of predominant first-reaction to a given tweet.

As mentioned in \cref{section:data}, the \retweet is extracted in a way to be as random as possible, since we do not consider the dynamic environment and rather only do content-based prediction.
For future work, it would be beneficial to extend the method by taking some prior knowledge into consideration. Time of posting, status quo when the tweet is posted, etc., can be influential factors.
Consequently, more task-specific data collection strategies will be explored to extend  \retweet.

\bibliographystyle{IEEEtran}
\bibliography{IEEEICSC}
\end{document}